# SEMI-SUPERVISED SEGMENTATION OF MITOCHONDRIA FROM ELECTRON MICROSCOPY IMAGES USING SPATIAL CONTINUITY

*Yunpeng Xiao[1,2], Youpeng Zhao[2], Ge Yang[1,2]*

[1]School of Artificial Intelligence, University of Chinese Academy of Sciences, Beijing, 100049, China
[2]National Laboratory of Pattern Recognition, Institute of Automation, Chinese Academy of Sciences, Beijing, 100190, China

**ABSTRACT**

Morphology of mitochondria plays critical roles in mediating their physiological functions. Accurate segmentation of mitochondria from 3D electron microscopy (EM) images is essential to quantitative characterization of their morphology at the nanometer scale. Fully supervised deep learning models developed for this task achieve excellent performance but require substantial amounts of annotated data for training. However, manual annotation of EM images is laborious and time-consuming because of their large volumes, limited contrast, and low signal-to-noise ratios (SNRs). To overcome this challenge, we propose a semi-supervised deep learning model that segments mitochondria by leveraging the spatial continuity of their structural, morphological, and contextual information in both labeled and unlabeled images. We use random piecewise affine transformation to synthesize comprehensive and realistic mitochondrial morphology for augmentation of training data. Experiments on the EPFL dataset show that our model achieves performance similar as that of state-of-the-art fully supervised models but requires only ~20% of their annotated training data. Our semi-supervised model is versatile and can also accurately segment other spatially continuous structures from EM images. Data and code of this study are openly accessible at https://github.com/cbmi-group/MPP.

*Index Terms*—Image segmentation, semi-supervised learning, deep learning, mitochondria, electron microscopy

## 1. INTRODUCTION

Mitochondria are intracellular organelles that serve many essential physiological functions in eukaryotic cells [5]. Their morphology is critical in mediating their physiological functions. Abnormal mitochondrial morphology has been implicated in many human diseases, including cancer and neurodegenerative diseases [5]. Electron microscopy (EM) provides a powerful tool to visualize mitochondrial morphology at nanometer resolutions over large volumes. However, quantitative characterization of mitochondrial morphology requires accurate image segmentation. So far, fully supervised deep learning models provide overall the best segmentation performance but require substantial amounts of labeled data for training. However, the large volumes, limited contrast and low signal-to-noise ratios (SNRs) of EM images make manual annotation laborious, time-consuming, and prone to human errors. The diverse and complex morphology of mitochondria in 3D also poses a substantial challenge to manual annotation. To overcome these technical challenges, it is desirable to develop models that can be trained with limited labeled data and effectively utilize information from unlabeled data.

Various classical machine learning models have been used in previous studies. For example, Lucchi *et al*. [1] employed an approximate subgradient descent algorithm to minimize the margin-sensitive hinge loss in structured support vector machines (SVMs). Li *et al*. [8] used ridge detection to acquire mitochondrial membrane edges in a variational image segmentation model and optimized segmentation using group similarity information. Peng *et al*. [3] introduced a class of local patch patterns (LPPs) to encode contextual features to improve mitochondrial segmentation accuracy. Overall, these methods provide reasonable results but are limited in their generalization capacity and their ability to handle large EM datasets.

Deep neural networks (DNNs) [9], such as U-Net [2, 4] and V-Net [10], have been widely used for segmentation of 3D biomedical images. Oztel *et al*. [11] proposed to use a fully convolutional network (FCN) to segment mitochondria and applied several post-processing procedures, such as 2D spurious detection filtering, boundary refinement and 3D filtering, to improve results. Yuan *et al*. [7] introduced a 3D multi-task network that adds auxiliary centerline detection to account for shape information of mitochondria under limited labeled data. However, large volumes of unlabeled data are unused in model training in these studies. Semi-supervised learning (SSL) models segment by leveraging both labeled and unlabeled data and are well suited for many biomedical image segmentation applications. Recent studies [12-14] have developed semi-supervised segmentation models by incorporating prior knowledge on properties such as shape, location, and context. Spatial priors [15] have also been utilized as a regularization term in fully supervised training. However, to our knowledge, semi-supervised segmentation has not been used on challengeing EM images.

In this paper, we propose a semi-supervised model for segmentation of mitochondria from EM images by utilizing spatial continuity of their structural, morphological, and contextual information. The main contributions of our work are as follows: (1) We have developed a new semi-supervised segmentation model that encodes spatial continuity information for segmentation of 3D mitochondrial morphology from EM images. It achieves performance similar as that of state-of-the-art fully supervised models but utilizes ~20% of their training data. (2) We have developed a new morphological post-processing procedure to use spatial continuity for segmentation refinement. (3) We propose a new strategy that utilizes random piecewise affine transformation to synthesize comprehensive and realistic mitochondrial morphology for training data augmentation.

## 2. METHOD

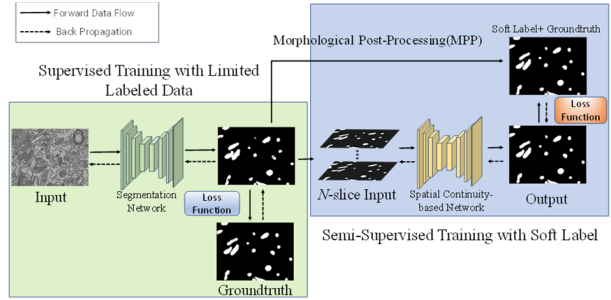

**Fig. 1** Architecture of the proposed model. It consists of a shape information sub-network $N_{se}$ with morphological post-processing (MPP) to refine coarse segmentation results and a spatial information sub-network $N_{sp}$ to encode continuous spatial information. Both sub-networks use a 2D U-Net architecture.

The overall architecture of our model is shown in Fig. 1. Its training and testing procedures are outlined in Algorithm 1. The objective is to utilize spatial continuity of structural, morphological, and contextual information of mitochondria from both labeled and unlabeled data for segmentation refinement.

---
*Algorithm 1*
*Training:*
**Data:** *Training image stack $S_{train}$. ~20% labeled slices are selected at equal intervals.*
**1.** *Train 2D segmentation network $N_{se}$ with the selected slices;*
**2.** *Use $N_{se}$ to segment all images in $S_{train}$ to get coarse segmentation stack $S_c$. Ground-truth is excluded from $S_c$;*
**3.** *Use morphological post-processing (MPP) with ground-truth to refine $S_c$ to get refined segmentation stack $S_f$, which includes both ground-truth and soft labels;*
**4.** *Train spatial continuity-based network $N_{sp}$. Use the coarse segmentation mask of the middle slice and adjacent N slices in $S_c$ as N-channel input, Output refined segmentation result of the middle slice. Use ground-truth or soft labels in $S_f$ as supervision information;*

*Testing:*
**Input:** *Test image stack $S_{test}$*
   *Trained segmentation network $N_{se}$*
   *Trained spatial continuity-based network $N_{sp}$*
**Output:** *Fine segmentation stack $S_f$ of $S_{test}$*
**1.** *Use $N_{se}$ to segment $S_{test}$ to get coarse segmentation stack $S_c$;*
**2.** *Use $N_{sp}$ to refine $S_c$ to get fine segmentation stack $S_f$;*

---

### 2.1 Supervised Training with Limited Labeled Data
A 3D stack of EM images can be viewed as a series of 2D slices. In a semi-supervised learning setting, we assume that only sparse slices in the training image stack are labeled and aim to segment an entire 3D stack of test images. There are different ways to select sparse slices from a 3D image stack for initial annotation. Continuous slices provide local contextual information but are less effective in providing global morphological information of mitochondria. We choose slices at equal intervals from a 3D image stack to balance local contextual and global morphological information.

### 2.2 Data Augmentation Using Piecewise Affine Transformation
Random flipping and rotation are two commonly used strategies for augmentation of natural images. Elastic deformation[16] is a commonly used augmentation technique to simulate morphological changes in EM images [2]. But as pointed out in [17], it may disrupt edge smoothness of mitochondria. In contrast, affine transformation can augment mitochondrial images while maintaining edge smoothness. Considering the diverse shapes of mitochondria, we find it suitable to use random piecewise affine transformation to synthesize comprehensive and realistic mitochondrial morphology from limited labeled data. Specifically, we use the piecewise affine function implemented in the *imgaug* package [18]. Detailed information on the piecewise affine transformation and performance comparison of elastic deformation versus piecewise affine transformation are presented in the Supplementary Material.

### 2.3 Supervised Segmentation Model with Refinement
When trained with limited labeled data, 2D U-Net usually generate inaccurate segmentation masks that contain many false positives (FPs) and false negatives (FNs). Previous studies improve segmentation results through multi-task learning [7]. However, it requires additional network parameters at high training cost and ignores essential latent information in unlabeled data. To alleviate this problem, we propose to use morphological post-processing to refine coarse segmentation obtained in the first stage.

**Morphological Post-Processing (MPP).** Several studies have proposed to use a series of post-processing steps such as filtering and boundary refinement to improve segmentation results [6, 11]. Here, we propose a new morphological post-processing (MPP) procedure to encode spatial continuity information from adjacent slices. It uses the following two operations:

***Foreground Union within Adjacent Slices***. An examination of ground truth of mitochondrial segmentation finds that foreground pixels in adjacent slices mostly overlap (Fig. S3) and that those pixels further away from edges are more likely to appear in adjacent slices. Therefore, performing morphological erosion on adjacent slices and taking its union with the segmentation result of this slice can effectively reduce false positives. The operation is described as follows:

$$A'_i = A_i \cup \Theta(A_j) \qquad (1)$$

where $A_i$ is the coarse label of the $i$-th slice and $A_i'$ is the soft label after processing, $A_j$ is the closest ground truth slice to $A_i$. $\Theta$ denotes a series of morphological erosion operations using a 3-by-3 kernel. The erosion is repeated ($n*|i-j|$) times on the image, where $n$ is a hyperparameter to be set for specific image stacks and $|i-j|$ denotes the distance from the current label to the closest ground truth slice. Generally, the lower the Z-axis resolution, the larger the $n$,

***Spatial Continuity-Based Refinement***. Under spatial continuity, pixels of the same coordinate between slices should not change drastically between foregrounds and backgrounds. If pixels of the same coordinate of adjacent slices are both in the foreground, the same coordinate pixels of the current slice are likely in the foreground. For each slice, to reduce false positives, the following formula is used:
$$A_i' = A_i \cap (A_{i-1} \cup A_{i+1}) \qquad (2)$$
And to reduce false negatives:
$$A_i' = A_i \cup (A_{i-1} \cap A_{i+1}) \qquad (3)$$
Here, $A_{i-1}$ and $A_{i+1}$ represent ground truth or coarse segmentation masks of adjacent slices. Further details are provided in Supplementary Material Fig. S4. With limited annotation data, when there are false positives or negatives in consecutive $n$ slices, we perform the above operations using a total of $n+2$ consecutive slices.

## 2.4 Semi-Supervised Segmentation Model Based on Spatial Continuity

The above MPP only refines the segmentation results from one dimension and does not fully consider the three-dimensional spatial information. This may lead to errors. However, manual design of 3D MPP is challenging. To address this problem, we train another network to further refine the segmentation results, using segmentation masks obtained after MPP as soft labels. Although the 2D U-Net provides a suitable architecture, it does not consider spatial information between slices. 3D U-Net incorporates rich spatial information from 3D input slices, but further experiments indicate that it has substantial bias towards middle slices. When a 3D image stack is input into a 3D U-Net, middle slices often have better segmentation results than edge slices (Supplementary Figure S2). This may be due to that edge slices can only use spatial information on one side, while middle slices can use spatial information from both sides.

Motivated by this observation, we design a spatial continuity-based model. Specifically, we use coarse segmentation masks of a middle slice and its adjacent slices to form an $N$-channel input, and we utilize segmentations masks after MPP as soft labels, which also contain ground truth labels. We train our network to predict the segmentation result of the middle slice.

When $N$ is 1, the spatial continuity-based network degenerates to a standard 2D U-Net. As $N$ increases, more adjacent slices are used as input so that more comprehensive spatial information can be obtained. However, spatial information from slices further away should not contribute much. Fig. 2 show that when $N$ is 15, our model achieves a balance between performance and model size. On the edge of the image stacks, we use nearest neighbor padding to ensure that edge slices can also get satisfactory segmentation results.

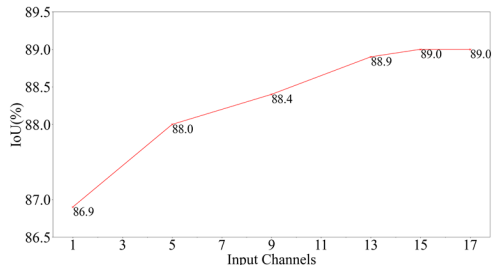

**Fig. 2** Performance of the spatial continuity-based sub-network with different input channel dimension, i.e. slice number $N$. The x-axis coordinates denote the number of different channels, i.e. slices; The y-axis coordinates denote Intersection over Union (IoU).

Unlike 3D U-Net, our spatial continuity-based model does not require splitting of images during training and testing. Furthermore, our model uses fewer parameters than 3D U-Net (7.7 M versus 16.3 M), making it easier to optimize. Other architectures may also be used in our spatial continuity-based model, as long as they take a stack of $N$ slices as their input and their output is of the same size as the individual slice.

## 3. EXPERIMENTS

We compare our method against feature-based [1, 3] and state-of-the-art fully supervised deep learning-based [2, 4, 6, 7] methods. We also conduct ablation experiments to examine the effectiveness of our proposed affine transformation, morphological post-processing (MPP) and spatial continuity-based sub-network. All experiments are conducted on a NVIDIA RTX 3090 GPU using the PyTorch framework.

### 3.1 Implementation Details

**EPFL Dataset.** We use the EPFL dataset [1], which consists of two 3D EM image stacks for training and testing, respectively. Each image stack has 165 slices, and the size of each slice is 768×1024. These images were acquired from the hippocampus of a mouse brain scanned by focus ion beam electron microscopy, and mitochondria are manually labeled by experts. Previous studies such as [6] used all the slices on the training dataset for training, while we used fewer than 20% of the slices for training.

**Training and Testing Settings.** For the EPFL dataset, we selected 32 labeled images from the original stack at the interval of one out of every five images. The selected images account for ~ 20% of the total number of labeled images. In comparison, fully supervised models are trained using all 165 labeled images. For the spatial continuity sub-network, we choose 15 consecutive slices as our input sequence. In MPP, the erosion hyperparameter $n$ is set to 3. In the training stage, we use random flip, rotation, and piecewise affine transformation for data augmentation. All models are optimized using RMSprop with a learning rate of 0.0005 and are trained for a total of 100 epochs. Batch size is set at 16. Dice loss is used for both sub-networks. In the testing stage, the average prediction of eight rotated images is used as model output.

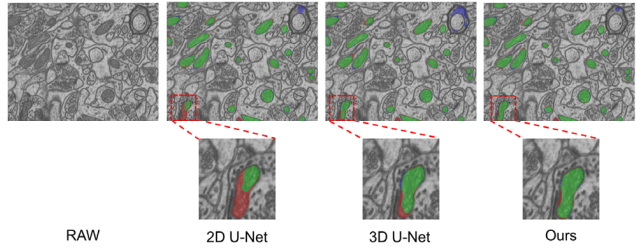

**Fig. 3** Comparison of segmentation results using different methods. Green: true positive (TP), Red: false negative (FN), and blue is false positive (FP).

### 3.2. Comparison with State-of-the-Art Methods

*Table 1.* Quantitative performance comparison of different methods for mitochondria segmentation on the EPFL dataset.

| Methods | Labels | Dice (%) | IoU (%) |
|---|---|---|---|
| Lucchi [1] | 165 | 86.7 | 75.7 |
| Peng [3] | 165 | 90.8 | 83.4 |
| 2D U-Net [2] | 165 | 91.4 | 84.4 |
| 3D U-Net [4] | 165 | 93.5 | 87.7 |
| Xiao [6] | 165 | 94.7 | 90.0 |
| Yuan [7] | 165 | **94.8** | **90.1** |
| **Ours** | **32** | 94.2 | 89.0 |

Table 1 summarizes a quantitative performance comparison of different methods on the EPFL dataset. Intersection over Union (IoU) and Dice (F1) similarity are used to evaluate segmentation performance. Consistent with previous findings, deep learning-based models outperform traditional feature-based methods. Our method achieves an accuracy of 89.0% in IoU and 94.2% in Dice, which are comparable to performance of fully supervised methods while using less than 20% of the labeled data. Qualitative results are shown in Fig. 3. Compared to 2D U-Net and 3D U-Net, our method produces more accurate segmentation results, with fewer FPs and FNs.

### 3.3. Ablation Experiments

We conducted additional experiments to examine the effectiveness of key components in our proposed model, including affine transformation, MPP and spatial-based model. As shown in Table 2, affine transformation substantially improves segmentation performance when labeled data is limited. MPP and spatial continuity-based model further enhance performance.

*Table 2.* Ablation experiment results on the EPFL dataset. 'Affine: affine transformation, 'MPP: morphological post-processing, SCM: spatial continuity model.

| Affine | MPP | SCM | Dice (%) | IoU (%) |
|---|---|---|---|---|
| - | - | - | 90.5 | 82.6 |
| ✓ | - | - | 92.5 | 86.9 |
| ✓ | ✓ | - | 93.6 | 88.0 |
| ✓ | ✓ | ✓ | 94.2 | 89.0 |

We also considered the effect of different numbers of labeled data on the segmentation results. As shown in Fig. 4, from 8 labels to 64 labels, segmentation metrics improve, and the best segmentation results have been largely achieved when using 32 labels.

### 3.4. Additional Experiments on EM Platelet Dataset

We further tested our method on the 3D EM platelet dataset [19]. The training image stack has 50 slices and the eval image stack has 24 slices. The size of each slice is 800× 800. There are seven categories in the label. For simplicity, we only segment the category of alpha granules. We set the erosion hyperparameter $n$ in MPP to 6. All other hyperparameters follow previous experiments. Detailed results are shown in Table S1. Together, the results confirmed the effectiveness of our method.

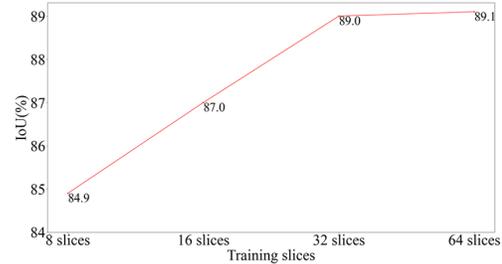

**Fig. 4** Performance of Using different number of training slices.

### 4. CONCLUSION

In this study we have developed a semi-supervised deep learning model that utilizes spatial continuity of structural, morphological, and contextual information to segment mitochondria from 3D EM images. Specifically, it uses morphological post-processing (MPP) to refine coarse segmentation results generated by a U-Net trained with sparse labeled data. Experimental results show that our method achieves similar performance as state-of-the-art fully supervised models but uses ~20% of their training data. Our study also has its limitations. First, our model only employs basic morphological operations to capture spatial continuity. There is still substantial room for improvement to derive more comprehensive priors through deeper analysis of mitochondrial morphology. Another limitation is that our 2D U-Net backbone uses mostly local visual information. Recent studies such as [20, 21] show that deep learning models that utilize global visual information provide better performance. These limitations will be addressed in our future work.

## 5. ACKNOWLEDGMENTS

This study was supported in part by the National Natural Science Foundation of China (grant 31971289, 91954201) and the Strategic Priority Research Program of the Chinese Academy of Sciences (grant XDB37040402).

## 6. COMPLIANCE WITH ETHICAL STANDARDS

This is a numerical simulation study for which no ethical approval was required.

# Supplementary Material

# Semi-Supervised Segmentation of Mitochondria from Electron Microscopy Images Using Spatial Continuity

## 1. Elastic Deformation versus Piecewise Affine Transformation

### 1.1 Elastic Deformation

First, generate random displacement fields $\Delta x$ and $\Delta y$, which represent the displacement of the pixel in the x-axis direction and the y-axis direction respectively. The size of $\Delta x$ and $\Delta y$ are the same as the image size. The displacement of each pixel is generated by the following formula:

$$\Delta x(x,y) = U \sim (-1,1) \tag{S.1}$$

$$\Delta y(x,y) = U \sim (-1,1) \tag{S.2}$$

Here, $\Delta x(x,y)$ is the displacement of the pixel at position $(x,y)$ in the x-axis direction. $\Delta y(x,y)$ is the displacement of the pixel in the y-axis direction. $U \sim (-1,1)$ is a uniform distribution from -1 to 1.

The original displacement field is transformed to the ultimate displacement field by the following formula:

$$\Delta x = \alpha * (G(\sigma_e) \otimes \Delta x) \tag{S.3}$$

$$\Delta y = \alpha * (G(\sigma_e) \otimes \Delta y) \tag{S.4}$$

Here, $G(\sigma_e)$ is the two-dimensional standard Gaussian kernel function and $\sigma_e$ is elastic coefficient. If $\sigma_e$ is large, the resulting values are very small because the random values average 0. $\alpha$ is a scaling factor that controls the intensity of the deformation. $\otimes$ represents convolution operation.

Then displacement fields are performed on each pixel:

$$(x', y') = (x + \Delta x(x,y), y + \Delta y(x,y)) \tag{S.5}$$

where $x'$ and $y'$ are the pixel coordinates after transformation. $x$ and $y$ represent the pixel coordinates of the original image. Through image interpolation, obtain the gray level value of $(x', y')$ coordinates in the original image, which is the new grey level value for pixel $(x,y)$.

### 1.2 Piecewise Affine Transformation

We first divide images to regions of sub-images by placing a regular grid of points on the original image. Then we randomly add translations to grids points to generate transformed sub-images. Translation of the grid points is based on the following formula:

$$Z_i = \lambda N \sim (0, \sigma_p) \tag{S.6}$$

$$(u', v') = (u + Z_{iu}, y + Z_{iv}) \tag{S.7}$$

Here, $Z_i$ is a Gaussian random variable with mean of 0 and standard deviation of $\sigma_p$. In our work, $\sigma_p$ is randomly set between 0.01 and 0.05. Step size $\lambda$ to be the same as the image size. $(u,v)$ refers to original coordinates and $(u', v')$ refers to coordinates after transformation.

Each sub-image division uses Delaunay's triangulation, and every three points determines a triangular sub-image. Each original sub-image and its corresponding transformed sub-image correspond to a set of affine transformation parameters, it is calculated by the following formula:

$$\begin{bmatrix} u'_i \\ v'_i \\ 1 \end{bmatrix} = \begin{bmatrix} a_1 & a_2 & c_1 \\ b_1 & b_2 & c_2 \\ 0 & 0 & 1 \end{bmatrix} \begin{bmatrix} u_i \\ v_i \\ 1 \end{bmatrix} (i = 1,2,3) \tag{S.8}$$

where $u'$ and $v'$ are the points coordinates after transformation. $u$ and $v$ represent the points original coordinates. Sub-image affine transformation parameters $a_1, a_2, b_1, b_2, c_1, c_2$ can be obtained by inserting the triangular points coordinates.

Then affine transformation is performed on pixels inside each sub-image:

$$\begin{bmatrix} x' \\ y' \\ 1 \end{bmatrix} = \begin{bmatrix} a_1 & a_2 & c_1 \\ b_1 & b_2 & c_2 \\ 0 & 0 & 1 \end{bmatrix} \begin{bmatrix} x \\ y \\ 1 \end{bmatrix} \quad (S.9)$$

where $x'$ and $y'$ are the pixel coordinates after transformation. $x$ and $y$ represent the pixel coordinates of the original sub-image. $a_1, a_2, b_1, b_2, c_1, c_2$ are parameters of affine transformation, which are calculated from the original sub-image and the deformed sub-image based on Equations S.6-S.8. Through image interpolation, obtain the gray level value of $(x', y')$ coordinates in the original image, which is the new grey level value for pixel $(x, y)$. These six parameters are the same for pixels inside the same sub-image to ensure edge smoothness of the mitochondria in the sub-image. They are usually different for different sub-images.

The more grid points, the image can be divided into more sub-images. However, it will increase the risk of the mitochondria being divided into different sub-images, which will disrupt the smoothness of the edges. Therefore, we use 2×2 grid points to divide the original image into two sub-images. The details are shown in Fig. S5.

**1.3 Comparison of Elastic Deformation versus Piecewise Affine Transformation**

Elastic Deformation performs a different displacement for each pixel, which ignores the global shape information of the image. As shown in Fig.S1 (c) and (d), when the ratio of $\alpha$ to $\sigma_e$ is large, smoothness of mitochondrial edges is disrupted; however, if the ratio is too small, there is almost no deformation. On the contrary, it can be seen from Fig.S1 (b) that the segmented affine transformation uses the same parameters to transform the pixels in the sub-image, which can ensure that the relative position of the mitochondrial edge pixels in the same sub-image does not change, therefore, it can generate large deformation while ensuring smoothness.

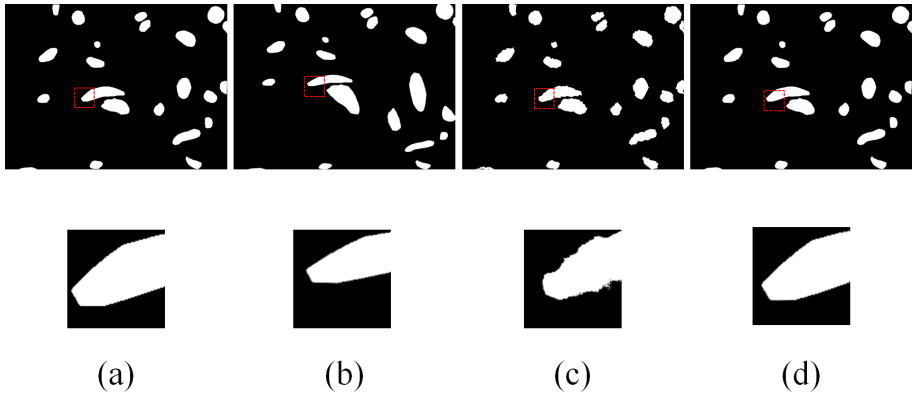

(a) (b) (c) (d)

**Fig. S1** Comparison of Elastic Deformation and Piecewise Affine Transformation. (a) is original label. (b) is label transformed by Piecewise Affine Transformation, $\sigma_p$ is set to 0.1. (c) is label transformed by Elastic Deformation, $\alpha$ is set to 40 and $\sigma_e$ is set to 5. (d) is also label transformed by Elastic Deformation, $\alpha$ is set to 40 and $\sigma_e$ is set to 20.

## 2. 3D U-Net Segmentation Results of the Same Slice at Different Depths

We verified the difference in the segmentation results of the same slice of 3D U-Net at different depths. The input of network is coarse labels and output is fine labels. The input depth is 16. The training dataset is the same as the spatial continuity-based network. Limited by memory, the input size is 16×256×256 in the training stage and 16×784×784 in the test stage. In test stage, we used a sliding window with a step size of 1 to generate input images on the test image stack and calculate the average of the Intersection over Union (IoU) of all 135 slices at 16 different depths except for the top and bottom 15 slices, a total of 30 slices, which cannot be used as input at every depth. We repeated the experiment three times in different areas of the test image. As shown in Fig.S2, it can be seen that middle slices usually have much better segmentation results than edge slices.

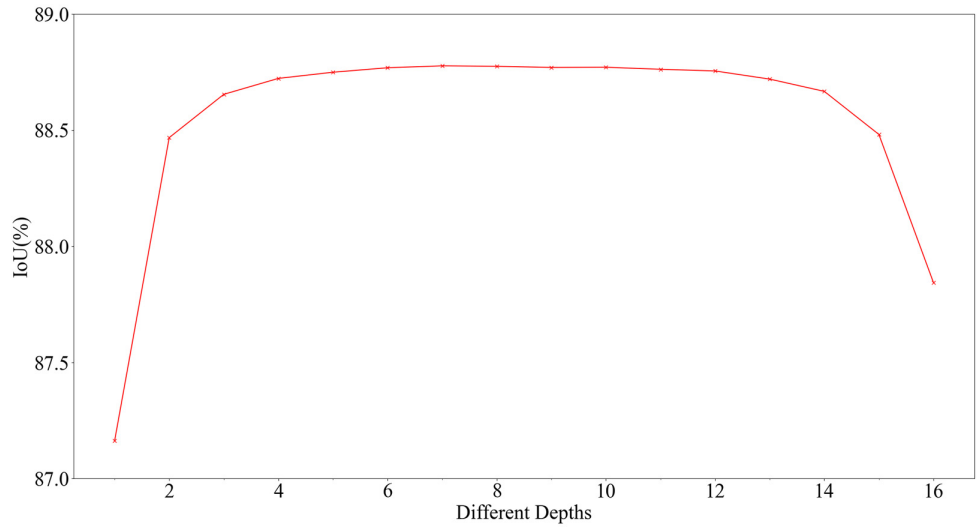

**Fig. S2** The average IoU of same slices at different Depths. Since only one side of the spatial information can be used, the edge slice segmentation result is significantly lower than the middle slice.

## 3. Other Supplementary Figures

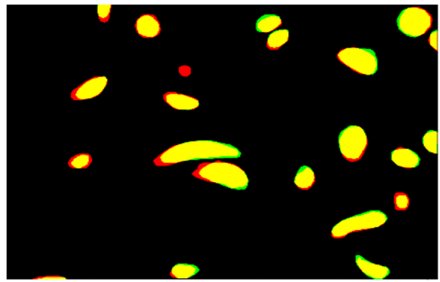

**Fig. S3** An example of morphological continuity. First and sixth slice ground truth mask in training dataset. Yellow: overlapping foreground of two image slices. Red: foreground only in first slice. Green: foreground only in sixth slice.

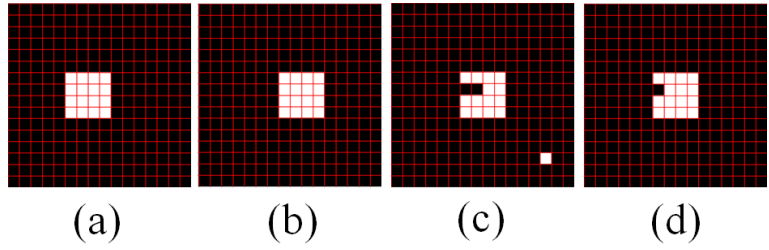

**Fig. S4** Spatial continuity-based segmentation refinement. (c) is the segmentation mask to be processed, (a) and (b) are masks of adjacent slices. (d) is the final mask obtained by after applying eq. 2 and 3.

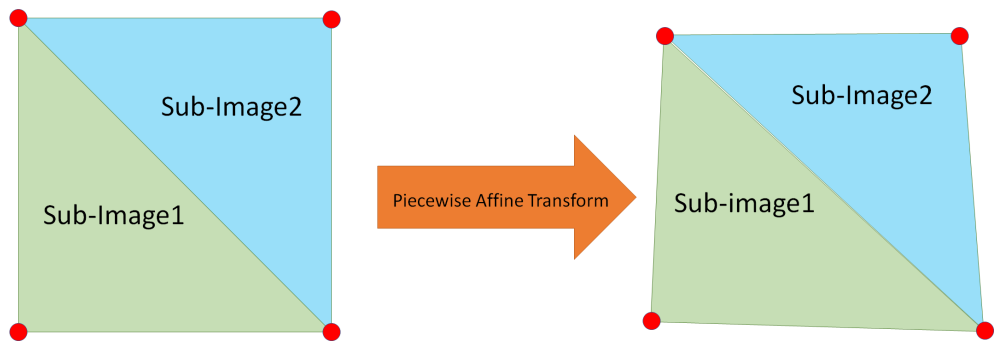

**Fig. S5** An example of 2×2 grid points Piecewise Affine Transformation.

*Table S1.* Quantitative comparison of different methods for alpha granule segmentation on 3D EM platelet dataset.

| Methods | Labels | Dice (%) | IoU (%) |
| --- | --- | --- | --- |
| 2D U-Net | 50 | 68.0 | 51.4 |
| 3D U-Net | 50 | 64.9 | 48.0 |
| **Ours** | **10** | **71.4** | **55.6** |

## 4. Morphological Post-Processing (MPP) Algorithm

---

**Algorithm 1:** Morphological post-processing

**Input**  : Coarse label of the image stack $S$ of size $d \times h \times w$,
$S = s_0, s_1, ... s_{d-1}$
Groundtruth $G = g_0, g_1, ..., g_j$,
Connected domain pixel number threshold $T_c$

**Output:** Binary image stack after morphological processing $S_B$

1  **for** $i \leftarrow 0$ **to** $d-1$ **do**
2      **if** $s_i$ *has its corresponding groundtruth* $g_k$ **then**
3          $s_i \leftarrow g_k$
4      **end**
5  **end**
6  **for** $i \leftarrow 0$ **to** $d-1$ **do**
7      **if** $s_i$ *has no its corresponding groundtruth* **then**
8          *Find nearest grountruth* $g_j$;
9          $c_j \leftarrow$ Corrde($g_j$);
10        $c_j \leftarrow$ Removeconnecteddomains($c_j$);
11        $s_i \leftarrow$ Union($(c_j, s_i)$);
12     **end**
13 **end**
14 **for** $i \leftarrow 0$ **to** $d-1$ **do**
15     **if** $s_i$ *has no its corresponding groundtruth* $g_i$ *and has two nearest groundtruth* $g_l$ *and* $g_k$ **then**
16        $s_i \leftarrow$ Intersection($s_i$,Union($g_l, g_k$));
17        $s_i \leftarrow$ Union($s_i$,Intersection($g_l, g_k$));
18     **end**
19 **end**
20 **for** $i \leftarrow 0$ **to** $d-1$ **do**
21     **if** $s_i$ *has no its corresponding groundtruth* $g_i$ *and has two nearest groundtruth or coarse label* $l_j$ *and* $l_k$ **then**
22        $s_i \leftarrow$ Intersection($s_i$,Union($l_j, l_k$));
23        $s_i \leftarrow$ Union($s_i$,Intersection($l_j, l_k$));
24     **end**
25 **end**
26 $S_B \leftarrow S$

---